\pdfoutput=1
\documentclass[11pt]{article}

\usepackage[final]{acl}

\usepackage{latexsym}
\usepackage{times}

\usepackage[utf8]{inputenc}

\usepackage{microtype}

\usepackage{inconsolata}

\usepackage{graphicx}
\usepackage{todonotes}
\usepackage{array}
\usepackage{booktabs}
\usepackage{multicol,multirow}
\usepackage[table-format=2.2, round-mode=places, round-precision=2]{siunitx}
\usepackage{tikz}
\usepackage{pgfplots}
\pgfplotsset{compat=1.18}
\usepackage[export]{adjustbox}
\usepackage{subcaption}
\usepackage[T2A,T1]{fontenc} % before babel!
\usepackage[russian,english]{babel}	

\usepackage{linguex}

\title{Adapting Definition Modeling for New Languages: \\ A Case Study on Belarusian}

\author{Daniela Kazakouskaya \And Timothee Mickus \\[1em]
  University of Helsinki \\
  \texttt{firstname.lastname@helsinki.fi} \\
  \And Janine Siewert}

\begin{document}
\maketitle
\begin{abstract}
Definition modeling, the task of generating new definitions for words in context, holds great prospect as a means to assist the work of lexicographers in documenting a broader variety of lects and languages, yet much remains to be done in order to assess how we can leverage pre-existing models for as-of-yet unsupported languages.
In this work, we focus on adapting existing models to Belarusian, for which we propose a novel dataset of 43,150 definitions. 
Our experiments demonstrate that adapting a definition modeling systems requires minimal amounts of data, but that there currently are gaps in what automatic metrics do capture.
\end{abstract}

\section{Introduction}

Dictionaries are invaluable resources. On a sociological level, it is fairly well documented that dictionaries are linked to cultural identity \citep{dollinger-2016-national}.
From the point of view of the NLP scientist, lexicographic data has historically proven very useful for tasks ranging from word sense disambiguation \citep{10.1145/318723.318728} to representation learning \citep{hill-etal-2016-learning-understand}.
On the other hand, lexicography is a complex enterprise: writing a dictionary from scratch is a time-consuming process, which often limits the number of languages, dialects and sociolects which can effectively be documented.

Definition modeling, the NLP task of generating definitions for words in context, is a promising direction to better support lexicographers in their work.
Definition modeling has grown as a field since the seminal work of \citet{noraset-etal-2017-definition}: we now have access to mature systems that can produce definitions automatically for English, Russian  and other languages \citep{kutuzov-etal-2024-enriching}.
A direction that remains to be explored is whether these available pretrained definition modeling systems can be leveraged for as-of-yet unsupported languages.
We take the Belarusian language as  the object of our case study.
Our main research question is to explore what is necessary to adapt a definition model to a new language --- are large amounts of data necessary? Do we need base models trained for similar languages?
To that end, we introduce a novel dataset of over 43,000 definitions for Belarusian, with which we demonstrate that a minimal amount of data is often sufficient to adapt to a novel language with reasonable performance.

This object of study also requires, as a complementary step, that we discuss how these systems should be evaluated. This has already been a point of inquiry in previous works --- e.g., \citet{bevilacqua-etal-2020-generationary} whereas \citet{segonne-mickus-2023-definition} conducted manual evaluation. 
Here, we contrast measurements from automatic and manual evaluation, and underscore current limitations in the evaluation of definition modeling.
We make our code and data available at \href{https://github.com/kozochkadaniela/tsbm}{\tt github.com/kozochkadaniela/tsbm}.

\section{Related works}

% first paper in the field
Definition modeling, initially introduced by \citet{noraset-etal-2017-definition}, is the NLP task that consists in generating definitions \citep{gardner2022definition}.
If the original formulation of \citeauthor{noraset-etal-2017-definition} involved static word embeddings as inputs, the field has since then shifted to contextualized definition modeling, where models are tasked to produce definitions for words in context \citep{gadetsky-etal-2018-conditional}.
% survey

% other use case, skippable
% 

% some stuff on multilinguality and definition modeling
The most common use-case for a definition modeling system is to create tools that facilitate the understanding of rare or technical words \citep{balachandran-etal-2018-learning,huang-etal-2021-definition,jhirad-etal-2023-evaluating,huang-etal-2022-understanding,10.1145/3677389.3702536}: the appearance of novel terminology, slang and neologisms outpaces often what lexicographers can handle manually.
Another application is to automatize and support efforts for language documentation 
\citep{bear-cook-2021-cross}.
As for this latter purpose, if efforts have been made towards studying definition modeling in multilingual contexts
\citep[e.g.,]{mickus-etal-2022-semeval,kutuzov-etal-2024-enriching}, or for languages other than English (ranging from Portuguese, \citealp{dimas-furtado-etal-2024-dore}, to Japanese, \citealp{huang-etal-2022-jade}), limited work has been devoted to cross-lingual transfer --- a step necessary if we want to re-purpose systems to low-resource contexts where they are needed.
% \citep{kabiri-cook-2020-evaluating}.

% In the present work, we focus on Russian as a potential source language for cross-lingual transfer. 
% This is a practically motivated choice, in that some effective models already exist for this language \citep{kutuzov-etal-2024-enriching}; furthermore, it has been an object of study in multiple prior works, including several shared tasks
% \citep{mickus-etal-2022-semeval,fedorova-etal-2024-definition,fedorova-etal-2024-axolotl24}.

% \todo[inline]{Timothee}

\section{Experimental setting}

Our overall approach is to (i) finetuning existing definition modeling systems for Belarusian, varying some key characteristics in their training, such as the amount of data they have access to and the base model we finetune; (ii) compare and contrast automatic metrics to the manual evaluation by a native Belarusian speaker, using a correlation analysis.

\subsection{Dataset}

We retrieve our data from the \textit{Skarnik} online Russian-Belarusian dictionary,\footnote{\url{https://www.skarnik.by}} originally based on the academic dictionary published by \citet{tsbm} % in 1953 \citep{tsbm} (edited by Ya. Kolas, K. Krapiva, and P. Hlebka), 
and subsequently revised and regularly updated. 
The dataset was obtained directly from an open-access repository provided by its maintainers.  To ensure the reliability and consistency of the data, additional preprocessing steps were applied. These included the removal of incorrect or misparsed entries, particularly words accompanied by unrelated example sentences. Words containing typographical errors or non-linguistic symbols were manually corrected. Additionally, several entries lacked explicit part-of-speech (POS) annotations or included only partial morphological information (e.g., gender, tense) without specifying the syntactic category. In such cases, full POS tags were added based on the available morphological information. Additionally, functional words (e.g., prepositions, conjunctions, determiners) were excluded from the dataset, and only content words were retained for analysis.  
% To evaluate how the model performs with different amounts of training data, we split the dataset into five parts: 1\%, 3\%, 10\%, 30\%, and 100\% of the total dataset.  

\begin{table}[t]
    \centering
    \resizebox{0.65\columnwidth}{!}
    { %\footnotesize
    \begin{tabular}{>{\bf\arraybackslash}l r r r}
    \toprule
    & \textbf{Train} & \textbf{Val.} & \textbf{Test} \\
    \midrule
         N. items      & 40105 &  1486 & 1159 \\
         N. glosses    & 40073 &  1485 & 1558 \\
         N. headwords  & 28203 &  1060 & 1062 \\
         N. homographs &  1879 &    70 &   71 \\
    \bottomrule
    \end{tabular}
    }
    \caption{TSBM dataset statistics. N.~items tracks the number of distinct instances (glosses and examples). N. homographs corresponds to the number of headwords with exact homographs in Russian.}
    \label{tab:stats tsbm}
\end{table}

% Furthermore, for every entry, we automatically annotate whether the headword has an exact homograph in Russian by considering whether an entry exists for this word in Russian in Wiktionary.\footnote{Data retrieved through \url{kaikki.org}.}
We then construct train, validation and test splits such that (i) headword types are only assigned to a single split, (ii) the proportion of Russian homographs is constant across splits and (iii) the train split contains at least 40K instances. 

\subsection{Models}
We finetune the Russian Definition Modeling system of \citet{kutuzov-etal-2024-enriching}, an \texttt{MT0-XL} model of 3.7B parameters fine-tuned on the CoDWoE dataset \citep{mickus-etal-2022-semeval}. 
Taking inspiration from \citeauthor{kutuzov-etal-2024-enriching}, inputs are formatted as in \ref{ipt}:

\ex. \texttt{[EXAMPLE]} \foreignlanguage{russian}{Что такое} \texttt{[HEADWORD]}? \label{ipt}

We use definition glosses as target outputs.
Our models are all trained on the TSBM data (cf. above), using subsets of logarithmically-spaced sizes, namely $100^{0/4}\%=1\%$, $100^{1/4}\%\approx3.16\%$, $100^{2/4}\%=10\%$, $100^{3/4}\%\approx31.62\%$, and $100^{4/4}\%=100\%$ of the available training data.
We train three models for each subset with fixed random seeds.
We furthermore report the performances of \citeauthor{kutuzov-etal-2024-enriching}'s (not re-trained) Russian Definition Modeling system as a baseline, which we refer to as training with $0\%$ of the data.
Lastly, to provide a better grasp as to the effects of language similarity on the performances we observe, we also duplicate our experiments using the two other \texttt{MT0-XL}--based models of \citeauthor{kutuzov-etal-2024-enriching}, designed for Norwegian and English. %, as supplementary baselines.

% \section{Analysis}
\subsection{Automatic metrics}

 We report performances obtained with default metrics commonly used in NLG: BLEU \citep{papineni-etal-2002-bleu,post-2018-call}, BERTScore \citep{Zhang*2020BERTScore:},\footnote{
\texttt{bert-base-multilingual-cased} \citep{devlin-etal-2019-bert}
}
BLEURT \citep{sellam-etal-2020-bleurt}, and chrF++ \citep{popovic-2015-chrf,post-2018-call}.

While BLEU assesses precision based on the number of exact matches in the candidate and the reference definition, BERTScore is more flexible as it does not compare the candidate and reference directly, but instead computes the similarity of their contextual embeddings. This makes it possible to recognize similar semantics despite different word use, which improves robustness against word swapping and leads to a higher overlap with human judgments \cite{Zhang*2020BERTScore:}. However, unlike BLEU, the usefulness of BERTScore depends on the quality of embeddings, which can be an issue in low-resource scenarios such as the one we are dealing with. 

The other two metrics are less frequently used for definition modeling, but offer interesting perspectives worth investigating.
The chrF++ metric of \citeauthor{popovic-2015-chrf} assesses overlaps of character spans --- which is useful to measure, given that generated definitions can rely on morphological  relationships \citep{segonne-mickus-2023-definition} and  that character-level information can prove beneficial \citep{noraset-etal-2017-definition}.
BLEURT, on the other hand, is a neural metric which is based on a small collection of variant models; the different existing models provide a tradeoff between computational costs and match with human assessments \citep{pu-etal-2021-learning}. 

%For the results/discussion: synonyms not always recognised (some examples @Daniela?)

\subsection{Manual evaluation}

% \todo[inline]{Daniela and Janine: defining criteria}
For the manual evaluation, we chose the criteria informativeness, fluency, and correct language and circularity. 

\paragraph{Fluency.} Fluency evaluates grammatical correctness, naturalness of phrasing %, correct language 
and basic semantic coherence, i.e., whether the sentence makes sense even if it does not fully capture the intended meaning. Outputs rated 1 are fully natural, grammatically correct and fluent. A score of 0.5 is assigned to outputs with minor grammatical issues (e.g., an unexpected \begin{otherlanguage*}{russian}я–е\end{otherlanguage*} alternation in the stem) or slightly unnatural phrasing. Outputs rated 0 exhibit clear grammatical errors, non-existent word forms, or constructions that are confusing or ungrammatical.

\paragraph{Informativeness.} Informativeness assesses how well the output conveys the intended meaning of the gloss. Outputs rated with a score of 1 are clear and accurate. A score of 0.5 is assigned to definitions that are too broad, incomplete, or only partially informative. A score of 0 reflects outputs that are semantically uninterpretable, even if the general topic is somewhat correct, or cases where the model lists several synonyms and some of them are wrong.

\paragraph{Circularity.} Circularity assesses the extent to which a model repeats the headword in its generated definition. A definition is considered fully circular if it includes the headword itself or one of its inflected forms. If the definition uses a derivational form of the headword, it is classified as partially circular. Definitions that do not contain the headword or any of its inflectional or derivational variants are labeled as not circular. This categorization helps assess whether the model can produce semantically informative paraphrases without relying on forms morphologically related to the headword.

 \section{Results \& discussion}

% \begin{figure}
%     \centering
%     \includegraphics[max width=0.9\columnwidth]{figs/normalized-vals.pdf}
        
%     \caption{Improvement of model performances as a function of data size.}
%     \label{fig:perfs}
% \end{figure}

\begin{table}[t]
    \centering
    \resizebox{0.9\columnwidth}{!}{
        \begin{tabular}{p{2cm}l *{6}{S}}
\toprule
\multirow{2}{*}{\textbf{Metric}} & \multirow{2}{*}{\textbf{Model}} & \multicolumn{6}{c}{\textbf{Data size}}  \\
 &  & {{\textbf{0\%}}} & {{\textbf{1\%}}} & {{\textbf{3\%}}} & {{\textbf{10\%}}} & {{\textbf{31\%}}} & {{\textbf{100\%}}} \\
\midrule
\multirow{3}{*}{BERTscore} 
 & EN & 63.042188 & 69.641158 & 70.524345 & 70.946695 & 71.494748 & 72.661952 \\
 & NO & 62.163268 & 70.017570 & 70.869016 & 71.134383 & 71.814485 & 72.818549 \\
 & RU & 63.277536 & 69.715988 & 70.607679 & 71.008747 & 71.669227 & 72.873240 \\
\midrule
\multirow{3}{*}{BLEU} 
 & EN & 4.037113 & 8.261732 & 10.142410 & 11.599395 & 12.578854 & 14.201563 \\
 & NO & 1.825115 & 8.310286 & 10.505721 & 11.723656 & 13.086132 & 14.312612 \\
 & RU & 4.658171 & 8.433592 & 10.551497 & 11.694774 & 12.650574 & 14.221221 \\
\midrule
\multirow{3}{*}{\parbox{2cm}{BLEURT \\ 20 D3}} 
 & EN & 8.614977 & 26.912661 & 28.553509 & 29.483152 & 31.055115 & 33.260375 \\
 & NO & 6.553888 & 26.748287 & 28.697271 & 29.600252 & 31.348313 & 33.352214 \\
 & RU & 11.738029 & 25.617756 & 28.599273 & 29.556002 & 31.127062 & 33.625039 \\
\midrule
\multirow{3}{*}{\parbox{2cm}{BLEURT \\ 20 D6}}  
 & EN & 8.128295 & 25.514107 & 27.753332 & 28.869388 & 30.409234 & 32.435369 \\
 & NO & 7.600255 & 25.485236 & 27.914962 & 29.206855 & 30.760322 & 32.682765 \\
 & RU & 13.177936 & 24.854802 & 27.926986 & 28.997891 & 30.381991 & 32.805168 \\
\midrule
\multirow{3}{*}{\parbox{2cm}{BLEURT \\ 20 D12}}  
 & EN & 9.255533 & 23.433912 & 25.565971 & 26.989069 & 28.346850 & 30.786810 \\
 & NO & 9.039648 & 23.453194 & 25.947866 & 27.272592 & 28.832425 & 31.019250 \\
 & RU & 13.396747 & 23.510047 & 25.711298 & 26.811292 & 28.347714 & 31.000040 \\
\midrule
\multirow{3}{*}{\parbox{2cm}{BLEURT \\ 20}}  
 & EN & 5.667192 & 24.540401 & 27.779558 & 29.299963 & 30.945872 & 33.859577 \\
 & NO & 6.592098 & 24.647459 & 28.021104 & 30.076734 & 31.713125 & 34.123120 \\
 & RU & 12.668867 & 25.099541 & 27.868789 & 29.480215 & 31.505360 & 34.235296 \\
\midrule
\multirow{3}{*}{chrF++} 
 & EN & 2.052219 & 14.252544 & 16.821416 & 18.400197 & 20.336493 & 22.662056 \\
 & NO & 0.760306 & 14.199587 & 16.683267 & 18.321512 & 20.486556 & 22.729313 \\
 & RU & 9.905947 & 14.036769 & 17.029114 & 18.406038 & 20.379329 & 22.972708 \\
\bottomrule
\end{tabular}
    }
    \caption{Overview of automatic metrics (average of 3 runs; all metrics in a 0--100 range). }
    \label{tab:automatic metric}
\end{table}

\paragraph{Automatic metrics.}
Corresponding performances are shown in Table \ref{tab:automatic metric}.
As is apparent, we observe higher scores for larger datasets. 
% Timothee: the sentence below is actually describing fig:perf, which I had to comment out for space
The progress is usually highly similar across all metrics: the average across all datasets is usually obtained with 10\% of the data; performances increase to +1 std. dev. above this average when using 100\% of the data; even 1\% of the data significantly mitigates the poor zero-shot performances of the base models.
Difference between base models are rarely significant outside of zero-shot conditions.

\begin{figure}[t]
    \centering
        \includegraphics[max width=0.9\columnwidth]{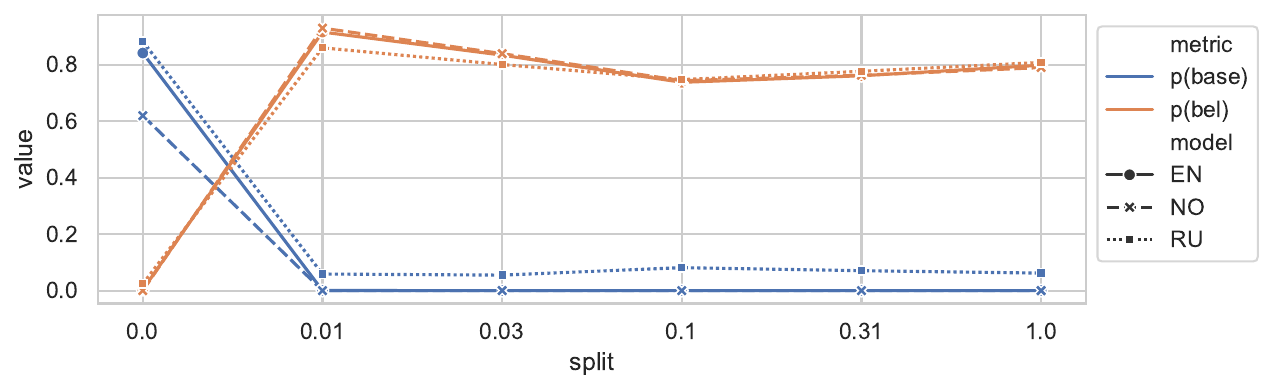}
    
    \caption{Language identification probability for Belarusian ($p$(\texttt{bel})) and base model language ($p$(base))}
    \label{fig:langprob}
\end{figure}

We also consider whether our models' outputs are indeed in Belarusian, or whether the base model being trained on another language impacts the output. 
We assess this using %the normalized probabilities derived from 
\texttt{langid.py} \citep{lui-baldwin-2012-langid}, in Figure \ref{fig:langprob}:
any amount of training data immediately gears all three models toward producing Belarusian, with a slight \emph{decrease} when using more than 1\% of the data as the model learn to produce more informative definitions.

It is worth remarking on the fact that metrics are surprisingly stable \emph{regardless of the language of the base model}. Performances with a Russian model re-trained for Belarusian are on par with what we observe with the Norwegian or English baselines.
This strongly suggests that adaptation does not depend on the similarity of the languages considered.
% \todo[inline]{Janine discussion}

% Table 2 values for English almost always lowest - is this statistically significant?

\paragraph{Manual analyses.}

% \todo[inline]{Daniela: tables for each criterion(?) with results per model}

\begin{table*}[ht]
\centering
%\resizebox{0.85\linewidth}{!}{
{\footnotesize 
\begin{tabular}{>{\bfseries}p{0.25cm}@{{~}} c p{1.5cm} p{   1.5cm}  p{2.5cm}  p{2.5 cm} p{2.5cm}  p{2.5cm} }
\toprule
 & \textbf{Score} & \textbf{Word} &  & \textbf{Target} & \textbf{Hypothesis} & \textbf{Translation} \\
\midrule
\multirow{8}{*}{\rotatebox{90}{\textbf{Fluency}}} & 1 & \begin{otherlanguage*}{russian}шанc\end{otherlanguage*} & chance & \begin{otherlanguage*}{russian}Умова, якая можа забяспечыць удачу, поспех; магчымасць.\end{otherlanguage*} & \begin{otherlanguage*}{russian} Шанц, магчымасць атрымаць што-н. \end{otherlanguage*} & A chance, an opportunity to get something. \\
& 0.5 & \begin{otherlanguage*}{russian}звечарэць\end{otherlanguage*} & to get dark & \begin{otherlanguage*}{russian}Пра надыход вечара, цемнаты.\end{otherlanguage*} &  \begin{otherlanguage*}{russian}Стаць вечарам.\end{otherlanguage*} & To become evening \\
& 0 & \begin{otherlanguage*}{russian}скляваць\end{otherlanguage*} & to peck & \begin{otherlanguage*}{russian}Клюючы, з'есці.\end{otherlanguage*} & \begin{otherlanguage*}{russian}Сцерці клявым.\end{otherlanguage*} & grammatically incorrect \\
\midrule
\multirow{13}{*}{\rotatebox{90}{\textbf{Informativeness}}} & 1 & \begin{otherlanguage*}{russian}берагчы\end{otherlanguage*} & to take care of & \begin{otherlanguage*}{russian}Клапатліва ахоўваць каго-, што-н., засцерагаць ад чаго-н.\end{otherlanguage*} & \begin{otherlanguage*}{russian} 
 Даглядаць, клапаціцца аб кім-, чым-н. \end{otherlanguage*} & to look after, to take care of someone or something. \\
& 0.5 & \begin{otherlanguage*}{russian}банкетка\end{otherlanguage*} & banquette (bench) & \begin{otherlanguage*}{russian}Тып нізкай табурэткі з мяккім сядзеннем прамавугольнай або круглай формы.\end{otherlanguage*} & \begin{otherlanguage*}{russian}Упрыгожванне на кухні, канапа, каб сесці.\end{otherlanguage*} & A decoration in the kitchen, a couch/sofa to sit on\\
& 0 & \begin{otherlanguage*}{russian}чаявыя\end{otherlanguage*} & tips & \begin{otherlanguage*}{russian}Грошы, якія даюць за дробныя паслугі, абслугоўванне.\end{otherlanguage*} & \begin{otherlanguage*}{russian}  У Беларусі — штраф, аплачваецца чаем. \end{otherlanguage*} & In Belarus, the fine is paid with tea  \\
\midrule
\multirow{12}{*}{\rotatebox{90}{\textbf{Circularity}}} & N & \begin{otherlanguage*}{russian}палігон\end{otherlanguage*} & military training area & \begin{otherlanguage*}{russian}Участак мясцовасці, спецыяльна абсталяваны для трэніровачнай стральбы і выпрабавання баявой тэхнікі.\end{otherlanguage*} & \begin{otherlanguage*}{russian}Група вайсковых часцей, якая мае пэўныя мэты.\end{otherlanguage*} & A formation of military units assigned to specific tasks \\
& P & \begin{otherlanguage*}{russian}дэбютаваць\end{otherlanguage*} & to debut & \begin{otherlanguage*}{russian}Упершыню выступіць на сцэне.\end{otherlanguage*} & \begin{otherlanguage*}{russian}Пачаць сваю дзейнасць, даць дэбют.\end{otherlanguage*} & To start one's career and make a debut \\
& F & \begin{otherlanguage*}{russian}вокладка\end{otherlanguage*} & book cover & \begin{otherlanguage*}{russian}Покрыўка кнігі, сшытка і пад.\end{otherlanguage*} & \begin{otherlanguage*}{russian}Тое, што і вокладка.\end{otherlanguage*} & Same as book cover\\
\bottomrule
\end{tabular}
}
%}
\caption{Examples illustrating annotation scores}
\label{tab:annotation-examples}
\end{table*}

% \paragraph{Fluency.}
For the manual analysis we examined 27 words with homographs in Russian and 50 without. We include examples of model productions for the criteria we annotate in Table~\ref{tab:annotation-examples}.

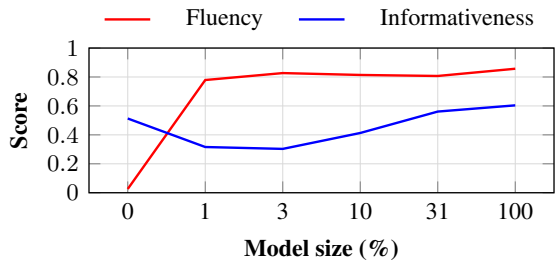
\begin{figure}[t]
\centering
\begin{tikzpicture}
\begin{axis}[
    width=1\linewidth,
    height=3.5cm,
    xlabel={\textbf{Model size (\%)}},
    ylabel={\textbf{Score}},
    xtick=data,
    xticklabels={0, 1, 3, 10, 31, 100},
    legend style={
        font=\small,
        cells={align=left},
        at={(0.5,1.05)}, anchor=south,
        draw=none,
        fill=none,
        column sep=1em,
        legend columns=2
    },
    ymin=0, ymax=1,
    ytick={0,0.2,...,1},
    grid=both,
    major grid style={line width=0.3pt,draw=gray!30},
    axis line style={line width=0.5pt},
    tick style={line width=0.3pt},
    tick label style={font=\small},
    label style={font=\small},
    every axis plot/.append style={mark=none, line width=0.9pt},
    cycle list name=color list
]

\addplot coordinates {
    (0, 0.026) (1, 0.779) (2, 0.827) (3, 0.814) (4, 0.807) (5, 0.857)
};
\addlegendentry{Fluency}

\addplot coordinates {
    (0, 0.513) (1, 0.316) (2, 0.303) (3, 0.413) (4, 0.561) (5, 0.604)
};
\addlegendentry{Informativeness}

\end{axis}
\end{tikzpicture}
\caption{Fluency and informativeness across data size.}
\label{fig:fluency_informativeness}
\end{figure}

A more global picture for fluency and informativeness is presented in Figure~\ref{fig:fluency_informativeness}.
Fluency remains consistently high across all data sizes. With only 1\% of the training data, the model already achieves a fluency score of 0.78, suggesting that it can produce natural and grammatically correct outputs even under low-resource conditions. Fluency slightly improves as more data become available, reaching 0.86 when the full dataset is used for fine-tuning. The proportion of Russian text in the retrained models doesn't exceed 2\%, and it typically appeared as either a single Russian word or the letter \begin{otherlanguage*}{Russian}и\end{otherlanguage*}. 
%
%\paragraph{Informativeness.} 
In contrast, the informativeness shows a more significant improvement as the amount of training data increases. Starting from a modest score of 0.32 in 1\%, informativeness increases to 0.60 when the entire dataset is used. This pattern highlights that, while fluency remains relatively stable even with limited training data, achieving accurate semantic alignment with the gloss requires larger datasets.

%\paragraph{Circularity.} 
As shown in Table~\ref{tab:circ}, full circularities decrease with model size, from 26\% when using 1\% of the data to 11\% when using all available data, indicating that larger models are more effective at avoiding circular definitions. Partial circularities remain consistently common across models, suggesting that models frequently reuse morphological forms of the target word, a strategy also used in human-written glosses \citep{segonne-mickus-2023-definition}. However, some predicted glosses, even from larger models, rely on morphological patterns and ultimately produced semantically incorrect meanings. Non-circular outputs are most frequent in the largest model (53\%), reflecting improved abstraction and lexical flexibility. Although we observe many non-circular outputs when using 1\% of the data, a large portion of them were semantically inaccurate.

A closer analysis of the errors reveals several recurring patterns. The model sometimes struggles with sense disambiguation, especially when the senses are closely related.  It may select the wrong meaning for polysemous or homographic entries. For example, \begin{otherlanguage*}{russian}{убяліць}\end{otherlanguage*}  means ‘to stain with something white’, but the model gives \begin{otherlanguage*}{russian}{пакрыць белым слоем}\end{otherlanguage*}(‘to cover with a white layer’), which is not exactly correct in context. Another issue is the generation of circular definitions, both full and partial. Furthermore, the model tends to prioritize morphological similarity over semantic accuracy, which can lead to incorrect output. For example, \begin{otherlanguage*}{russian}{банкаваць} \end{otherlanguage*}(`to act as a banker in a card game`) is misinterpreted as\begin{otherlanguage*}{russian} {уступаць у банк}\end{otherlanguage*} (`to join a bank`).

To conclude, the model often captures important features, such as verb aspect, argument structure, and correct word class. However, it still tends to produce incorrect or overly surface-level definitions. High-scoring outputs typically contain semantically appropriate expressions, but they do not fully capture all the nuances unlike human-produced definitions.

\begin{table}[t]
\centering
\resizebox{0.65\columnwidth}{!}
{
\begin{tabular}{l *{5}{S}}
\toprule
% \multicolumn{1}{c}{} & \multicolumn{5}{c}{\textbf{Model}} \\
% \cmidrule(lr){2-6}
 & {\textbf{1\%}} & {\textbf{3\%}} & {\textbf{10\%}} & {\textbf{31\%}} & {\textbf{100\%}} \\
\midrule
No \%     & 52.21 & 35.24 & 32.52 & 49.85 & 53.41 \\
Part \%   & 22.02 & 32.19 & 36.27 & 35.55 & 35.33 \\
Full \%   & 25.77 & 32.57 & 31.21 & 14.60 & 11.26 \\
\bottomrule
\end{tabular}}
\caption{Proportion of circular definitions}
\label{tab:circ}
\end{table}

% Full circularities decrease substantially with more training data. In 1\% and 3\% of the data, full circularities occur relatively frequently (approximately 9-10 instances on average), but this number drops to fewer than 3 instances with full data coverage. This pattern indicates that with sufficient training examples, the model moves away from simply replicating the headword in the definition.

\paragraph{Comparing manual and automatic assessments.}

\begin{table}[]
    \centering
    \resizebox{0.95\linewidth}{!}{
    \begin{tabular}{>{\bf}l@{~~} *{7}{S@{\quad}}}
\toprule
& {{\textbf{BERT-}}} & {{\multirow{2}{*}{\textbf{BLEU}}}} & \multicolumn{4}{c}{\textbf{BLEURT}} &   {{\textbf{chrF}}} \\
& {{\textbf{score}}} &  &   {{\textbf{D3}}}           &  {{\textbf{D6}}}          & {{\textbf{D12}}}           & {{\textbf{20}}}             &  {{\textbf{++}}} \\
\midrule
% Combined    &  26.681238 &  11.975414 &  36.575009 &  35.935036 &   39.408650 &    35.820533 &  39.876022 \\
Fluent      &  11.563806 &   6.599457 &  11.889287 &  13.634683 &   12.565080 &    10.914192 &   6.638126 \\
Informative &  25.525894 &  13.068272 &  34.257907 &  34.463912 &   39.791962 &    36.169501 &  40.375613 \\
\bottomrule
\end{tabular}
}
    \caption{Comparison of manual and automatic assessment using Spearman's $\rho$ ($\times 100$).}
    \label{tab:spearman metrics}
\end{table}

In Table~\ref{tab:spearman metrics}, we list coefficients of correlation between the automatic metrics and the manual annotation scores we detailed above.
We can note several key points:
Fluency is generally harder to capture than informativeness, with lower correlation scores; neural metrics such as BERTscore and BLEURT usually fare better than overlap metrics such as BLEU and chrF++.
Commonplace metrics in NLG in general and definition modeling in particular, such as BLEU and BERTscore, are in fact not the most suitable for definition modeling, especially when it comes to informativeness: in fact, chrF++ proves to be remarkably fit.
Lastly, what works for other NLG subfields need not apply in definition modeling contexts: while \citet{pu-etal-2021-learning} find BLEURT 20 to be a better model of human preferences than all of its distilled variants, here, BLEURT 20 D12 captures informativeness more appropriately, while BLEURT D6 is more appropriate as a model of fluency.

\section{Conclusions}
In this paper, we have studied how to adapt existing definition modeling systems to Belarusian.

To that end, we introduce a large dataset of Belarusian definitions and conduct extensive experimentation.
Small datasets can already achieve some success: even 1\% of the data collected was sufficient to ensure the generated definitions would be in Belarusian with a reasonably high degree of fluency.
Other characteristics often benefit from more data --- e.g., informative, non-circular definitions are more frequent in models trained on larger datasets. 
%It is also worth noting that we notice at best linear improvements over logarithmically-spaced dataset sizes: i.e., we need orders of magnitude more data to significantly improve performances.

Lastly, further research is necessary in order to properly automatize the assessment the quality of generated definitions: metric rankings from previous work do not translate to definition modeling in Belarusian; none of the metrics we tested capture fluency; and metrics can very greatly in their ability to describe informativeness.

% \section*{Limitations}

\section*{Acknowledgments}

This work is supported by the Research Council of Finland through projects No.~342859 ``CorCoDial -- Corpus-based computational dialectology'' and No~353164. ``Green NLP -- controlling the carbon footprint in sustainable language technology.''

\bibliography{anthology,custom}

\appendix

\end{document}